\documentclass[letterpaper]{article} 
\usepackage{aaai25}  
\usepackage{times}  
\usepackage{helvet}  
\usepackage{courier}  
\usepackage[hyphens]{url}  
\usepackage{graphicx} 
\usepackage{natbib}  
\usepackage{caption} 
\usepackage{algorithm}
\usepackage{algorithmic}

\usepackage{newfloat}
\usepackage{listings}
\DeclareCaptionStyle{ruled}{labelfont=normalfont,labelsep=colon,strut=off} 
\floatstyle{ruled}
\newfloat{listing}{tb}{lst}{}
\floatname{listing}{Listing}
\title{TechOps: Technical Documentation Templates for the AI Act}

\author{
    Laura Lucaj\textsuperscript{\rm 1,2},
    Alex Loosley\textsuperscript{\rm 3,4},
    Håkan Jonsson\textsuperscript{\rm 4},
    Urs Gasser\textsuperscript{\rm 2},
    Patrick van der Smagt\textsuperscript{\rm 5,6}
}
\affiliations {
    \textsuperscript{\rm 1}Volkswagen Group,\hspace{5mm}
    \textsuperscript{\rm 2}Technical University of Munich,\\
    \textsuperscript{\rm 3}Capgemini Deutschland GmbH,\hspace{5mm}
    \textsuperscript{\rm 4}Zalando SE,\\
    \textsuperscript{\rm 5}Foundation Robotics, \hspace{5mm}
    \textsuperscript{\rm 6}ELTE University Budapest\\
    \texttt{laura.lucaj@argmax.org}
}

\usepackage{bibentry}

\nocopyright

\begin{document}

\maketitle

\begin{abstract}
Operationalizing the EU AI Act requires clear technical documentation to ensure AI systems are transparent, traceable, and accountable. Existing documentation templates for AI systems do not fully cover the entire AI lifecycle while meeting the technical documentation requirements of the AI Act.

This paper addresses those shortcomings by introducing open-source templates and examples for documenting data, models, and applications to provide sufficient documentation for certifying compliance with the AI Act. These templates track the system’s status over the entire AI lifecycle, ensuring traceability, reproducibility, and compliance with the AI Act. They also promote discoverability and collaboration, reduce risks, and align with best practices in AI documentation and governance.
  
The templates are evaluated and refined based on user feedback to enable insights into their usability and implementability. We then validate the approach on real-world scenarios, providing examples that further guide their implementation: the data template is followed to document a skin tones dataset created to support fairness evaluations of downstream computer vision models and human-centric applications; the model template is followed to document a neural network for segmenting human silhouettes in photos. The application template is tested on a system deployed for construction site safety using real-time video analytics and sensor data. Our results show that TechOps can serve as a practical tool to enable oversight for regulatory compliance and responsible AI development.

\end{abstract}

%

\section{Introduction}
AI is increasingly used in areas where managing its risks is difficult, especially when protecting people’s fundamental rights~\cite{brundage2019responsible,o2017weapons,barocas2016big,critch2023tasra,solow2023algorithmic}. Many AI systems use complex, opaque models that make it hard to explain their predictions and outcomes. This complexity makes it challenging to manage risks throughout design, development, and deployment \cite{mittelstadt2016ethics,smuha2021beyond}.

Policies are emerging worldwide to address the harms and risks associated with the deployment of AI systems. The UK \cite{UKAIRegulations2024} and Switzerland \cite{SwissAIPositionPaper2021} have been adapting existing laws to govern AI. In contrast, globally, approaches vary from strict frameworks in China \cite{ChinaAIRegulation2022, ChinaAINorms2021} and Brazil \cite{BrazilAIBill2023} to sectoral or executive orders and proposed bills in the US \cite{USTrustworthyAI2020,USAILeadership2019,USBillS2551_2021,USAIBillOfRights2022} with stricter state-level legislations such as Colorado and many other federal states \cite{ColoradoAIAct2024}.
The EU AI Act is the first law in the world to comprehensively regulate AI systems, categorizing them into different risk categories to determine the legal requirements they are subject to~\cite{EU_AI_Act_2024}.

Nonetheless, turning such abstract legal requirements into operational solutions remains a challenge.
The debate has focused on methods and tools to solve such gaps and the critical need for mechanisms that enable adequate oversight of AI systems, such as robust documentation~\cite{lucaj2023ai,veale2021demystifying,alder2024ai,diaz2023connecting}.
Documenting an AI system enables transparency, accountability, as well as proof of adherence to legal requirements \cite{gebru_datasheets_2021,mitchell_model_2019,chmielinski2024clear,kroll2021outlining,raji2020closing,naja2021semantic}.

Prior work has already provided several artifacts to support the recording of specific lifecycle processes or components, or to provide a high-level mapping of the AI Act’s technical documentation requirements. 
However, none of these tools offers comprehensive coverage of the whole AI system lifecycle in a manner that meets the AI Act's obligations while also enabling a coherent and actionable overview of system functionality. 
To bridge this gap, we introduce TechOps, a set of open-source automatable templates for documenting data, models, and applications.  
Our templates are designed to fully align with the AI Act’s technical documentation requirements by following all stages of the AI lifecycle. This approach facilitates a clearer understanding of system performance, supports the timely identification and resolution of issues, improves overall quality, and helps to prevent the accumulation of technical debt.

Currently, there is a critical need to develop comprehensive technical documentation templates that account for all the processes throughout the AI lifecycle. 
This work bridges best‑practice AI documentation and the EU AI Act’s practical requirements.
In the next section, we outline the AI Act’s documentation requirements and current best practices. We then introduce TechOps, delineate the structure of each template, and explain how they enable tracking the system’s status over the entire AI lifecycle, ensuring traceability, reproducibility, and compliance with the AI Act. We then evaluate the templates on user feedback. This helps us identify implementability and usability issues, as well as opportunities. 
We illustrate this with real-world datasets, models, and applications, then compare TechOps with current AI documentation practices, summarising feedback from industry stakeholders. Finally, we outline its limitations and propose future improvements.

\section{Related Work}

Documentation is key to AI transparency and accountability. It records decisions across the lifecycle---such as system purpose, design choices, and development steps---so stakeholders can assess behavior, limitations, and risks \cite{gebru_datasheets_2021, mitchell_model_2019, konigstorfer2022ai, winecoff2024improving}. Structured documentation also supports audits, enabling regulatory oversight and responsible deployment \cite{arnold2019factsheets, birkstedt2023ai, arnold2024documentation}. Here, we position our work within existing literature, underscoring the role of structured documentation for compliance, accountability, and transparency.

\subsection{Documentation and AI Act Compliance}
The EU AI Act sets out detailed technical documentation requirements (Art.\ 11, 56, Annex IV) to support transparency, accountability, and demonstrable compliance \cite{EU_AI_Act_2024}. These include a description of the system’s intended purpose and interactions, along with comprehensive records of its design, data use, testing procedures, performance metrics, risk management strategies, and post-market monitoring plans. The requirements are summarised in Fig.~\ref{fig:templateAIActmapping}. 

Studies have identified the necessary information to be included when documenting AI systems for AI compliance \cite{hupont2023documenting,golpayegani2024ai}.  
Two methods specifically target AI Act compliance: Use Case Cards \cite{hupont2024use}, which document intended purpose and stakeholders, and AI Cards \cite{golpayegani2024ai}, a high-level template covering technical features and risk management. While valuable, these approaches do not capture the full complexity of AI system components or the technical practices across the entire machine learning lifecycle, design, development, deployment, and post‑market monitoring. They lack the granularity needed for a complete understanding of such systems. In the next section, we review best practices in AI documentation and show how our work addresses these gaps to operationalize the AI Act’s technical documentation requirements while retaining established practices.
The European Commission recently published a concise documentation template for general-purpose AI models to provide an overview of the data origin, list main data collections, and explain other sources used \cite{GPAI2025}.
While the current documentation guidance for GPAI focuses primarily on transparency around training data sources and rights reservations, TechOps extends this scope by incorporating explicit fields for disclosing user data usage, data protection impact assessments, and aligning documentation practices with core GDPR obligations often overlooked in existing guidance.

\subsection{Existing documentation practices}
Early oversight is critical, as inadequate governance can lead to costly retrofitting or fines when regulations take effect~\cite{holland2020dataset}. Data documentation has emerged as a foundational oversight tool. "Datasheets for Datasets"~\cite{gebru_datasheets_2021} set a de facto standard by outlining questions on purpose, composition, processes, distribution, maintenance, and impact. Other templates---such as Open Datasheets~\cite{roman2023open}, Dataset Nutrition Labels~\cite{holland2020dataset,chmielinski2022dataset}, and full lifecycle frameworks~\cite{hutchinson2021towards}---extend this work to improve usability, transparency, and governance. Drawing from software engineering, some frameworks document the dataset lifecycle from requirements through maintenance~\cite{hutchinson2021towards}. Tools like Data Cards~\cite{pushkarna2022data}, Data Statements~\cite{bender2018data}, and Data Portraits~\cite{marone2024data} expose decisions affecting model performance, bias, and exclusion. Domain-specific approaches---Augmented Datasheets~\cite{papakyriakopoulos2023augmented}, Reflexive Documentation~\cite{miceli2021documenting}, and CrowdWorkSheets~\cite{diaz2022crowdworksheets} target accountability in areas such as speech and vision.

Model documentation practices have evolved alongside system-level approaches to give a full view of a model’s design and development. Properly documenting the model lifecycle ensures it is not only functional but also reliable, fair, and aligned with its intended purpose.

Early adoption is key to assessing AI systems and prompting developers to identify, understand, and address product limits before release~\cite{raji2022fallacy}. Model Cards~\cite{mitchell_model_2019} provide a template to document an AI model’s intended use, performance, and limitations. They concisely report performance across domains to help prevent unintended uses and unsuitable applications.
Inspired by ``Supplier's Declarations of Conformity'' (SDoCs),
Factsheets \cite{arnold2019factsheets} expand this to include provenance, safety, and security considerations. Related frameworks include Method Cards \cite{adkins2022method}, System Cards \cite{procope2022system}, and Explainability Fact Sheets \cite{sokol2020explainability}, each targeting different issues such as detection of functional fallacies, transparency in the architecture, and explainability. Moreover, tools like Value Cards \cite{shen2021value} and Interactive Model Cards \cite{crisan2022interactive} further explore trade-offs and promote the inspection of specific aspects of the ML lifecycle, and understand the functionality and limitations of the systems. While these approaches represent significant contributions, neither fully captures the technical documentation requirements of the AI Act and the intricacies of AI system components or the technical practices applied across the AI lifecycle, including design, development, deployment, and post-market monitoring. This limitation makes it difficult for large organizations with diverse stakeholders to develop a comprehensive understanding of these systems.
Existing studies adopt varying perspectives on documentation. Some propose static documents describing the dataset or model at a specific point in time, while others advocate for evolving documents or interactive dashboards.  However, it is important to document the status of the AI System at every point in time separately to make sure that everything can be tracked back, and past results can be easily reconstructed \cite{etami2023}.
Hence, a specific version of the documentation templates should be treated as an immutable artifact at a certain point in time.
Therefore, to address such a gap whilst maintaining the existing best practices, we introduce three separate templates for data, models, and applications to enable compliance with the requirements of the AI Act, as well as enabling sustainable maintenance of such practices within organizations and relevant stakeholders.

\section{TechOps Development Methodology}

\begin{figure}
	\vspace{0mm}
	\centering
	\includegraphics[width=0.9\columnwidth]{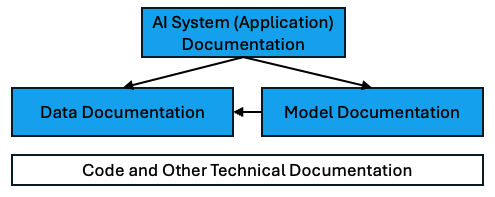}
        \caption{\textbf{AI system documentation hierarchy.} Arrows point in the direction of referral. For example, applications that rely on various models and datasets should refer to the corresponding documentation. Going the opposite direction, model and dataset documentation can also refer to known applications but do not have to.}
	\label{fig:documentation-hierarchy}
\end{figure}

To align with established best practices in AI documentation, this paper introduces three distinct documentation templates: one for data, another for models, and a third for applications. This separation acknowledges the distinct roles each component fulfills, letting those responsible for each component focus on providing documentation they are responsible for, while referencing information from the other components, so that information is not duplicated (see Fig.~\ref{fig:documentation-hierarchy}).
The documentation templates presented in this paper merge the technical documentation requirements of the AI Act with the documentation of the different stages of the ML lifecycle and the practices of machine learning operation (MLOps). MLOps streamline all the practices conducted throughout the lifecycle of AI systems, from design to continuous integration/delivery, data management, model development, testing and validation, deployment (including cloud and edge), post-market monitoring, and continual learning. Therefore, MLOps provide a guide for implementing all documentation requirements throughout the design, development, and deployment lifecycle of AI systems \cite{billeter2024mlops}.

The templates were developed iteratively, incorporating feedback from diverse stakeholders across the AI lifecycle and from different companies (see Tab.~\ref{tab:interview}). Participants included AI system developers, data and model platform engineers, tech law and governance experts, managers, and non-technical product specialists. A key challenge was balancing the right level of technical, legal, and ethical detail with readability. Because documentation serves multiple audiences, it had to reflect their varied responsibilities and backgrounds. Many existing templates sacrifice technical depth for simplicity; our approach preserves necessary detail on processes that link system performance to legal and ethical considerations. We focused on balancing conciseness with the detail needed to prove compliance and give all relevant stakeholders a clear view of design, development, and deployment. To help maintain both overview and depth, we recommend rendering documentation so users can navigate high-level summaries and easily drill down into specific sections (tested method described below).

These documentation artifacts aim to help all responsible stakeholders for each AI lifecycle component understand what must be documented for AI Act compliance and to assess the system’s quality, reliability, and robustness.

By following best practices and dividing the approach into three templates, we created a maintainable solution for organizations. Each stakeholder can focus on the sections relevant to their role: data teams on provenance, collection, preprocessing, labeling, and curation; model teams on training configurations, evaluation metrics, hyperparameters, and performance (while referencing data documentation); and application teams on intended use, limitations, and user interaction (while referencing model and data documentation). This separation lets stakeholders access only relevant details, improving clarity, transparency, accountability, and compliance across the lifecycle while supporting effective oversight and evaluation of each component.

One of the main challenges was the translation of abstract legal and ethical requirements into concrete documentation of the technical measures implemented across the AI system lifecycle---spanning design, development, deployment, and post-market monitoring~\cite{shneiderman2020bridging}. The templates we propose address this issue by blending legal and technical language to enable usability across all stakeholders. To reflect the diversity of roles involved in AI development, the templates accommodate input from multiple stakeholders, each responsible for distinct phases and tools.
For example, technical documentation of data management processes is directly linked to broader questions about their ethical and regulatory implications. This traceability enables stakeholders to assess both the localized and systemic impact of each process, offering granular insight as well as a holistic view of compliance.
Another challenge was making the templates generalizable across organizational functions with different terminology, tools, and data formats. We addressed this by unifying role-specific documentation requirements into a single structure, enabling contributors to add relevant information without losing completeness. For highly technical phases, the templates provide clear definitions and step-by-step guidance so governance stakeholders---including auditors, AI testers, and regulators---can access and understand the content, promoting transparency and accountability across the lifecycle.

\section{TechOps}

\begin{figure*}[t]
	
    \centering
    \includegraphics[width=1.0\textwidth]{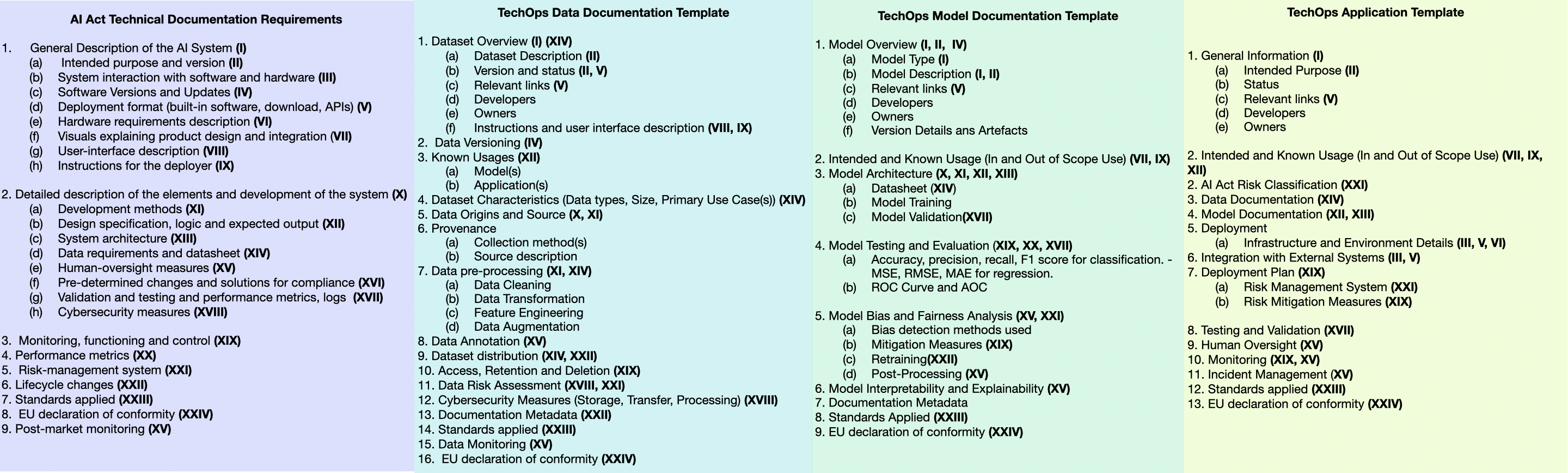}
    \caption{\textbf{AI Act Requirements and Template Mapping.}}
    \label{fig:templateAIActmapping}
	
\end{figure*}

We introduce the TechOps documentation templates, divided into data, model, and application templates, for all processes along the AI lifecycle that must be documented to enable the necessary oversight for compliance with the AI Act (see Fig.~\ref{fig:templateAIActmapping}).
TechOps can enable organisations to delineate design choices and development processes, as well as monitoring the performance after market deployment, and document roles and responsibilities along the AI lifecycle.

\subsection{Data Documentation Template}
This template builds on existing data documentation literature \cite{gebru_datasheets_2021,roman2023open,pushkarna2022data,chmielinski2022dataset,holland2020dataset,bender2018data}. It also complements existing efforts in data documentation, fostering transparency regarding the origin, quality, and potential biases of the data~\cite{hutchinson2021towards}. Moreover, this section builds on the work assessing the quality dimensions of data that must be documented for transparency and explainability~\cite{afzal2021data,castelijns2020abc,hiniduma2024ai}. That includes greater transparency about data and accountability for decisions made when developing it~\cite{kale2023provenance,kroll2021outlining}. In this paper, we introduce a framework to enable compliance as well as transparency and accountability throughout the development of the dataset. 
This approach enables organizations to systematically record all critical stages of the data governance lifecycle, facilitating the evaluation of data quality metrics, assessment of model performance, and adherence to the legal requirements of the AI Act \cite{bhardwaj2024machine, bhardwaj2024state,micheli2023landscape, diaz2023connecting}.
We based the design of this artifact on the studies that guide the design of templates based on ML practitioners' needs, such as integration into existing tools and workflows~\cite{heger2022understanding}. 

This template focuses on the requirements for data governance laid out in Article 10 and Article 11 paragraph 2(d) \cite{EU_AI_Act_2024} for documenting datasets, including data types and characteristics, origin and source, preprocessing steps, transformation steps that enable inferences about the quality as well as the annotation and validation steps to detect implicit data biases and limitations (i.e.\ what biases existed in the data generating process?)
Such documentation captures essential metrics for compliance and overall data quality, such as accuracy, completeness, consistency, timeliness, and fairness.
This approach enables the overview of the composition of the dataset, ensures transparency by helping downstream dataset users (e.g.\ developing an AI Model or evaluating an AI System), understand the potential limitations of the data selected, and whether it is appropriate for the intended use.

It is important to note that datasets, though often curated with a particular AI Model or AI System in mind, are not necessarily associated with any one AI Model or AI System.
Thus, the data documentation template is designed to be filled without complete knowledge of all downstream AI models and systems. The focus is on enabling data owners to make clear statements about the intended and unintended downstream usages of the dataset, potentially also with hypothetical examples of appropriate and inappropriate AI models and systems.
As usages become known, downstream developers that use the dataset are encouraged to contribute information to a ``Known Usages'' section.

As a generic example portraying the  version-controlled, distributed approach to contributing to dataset documentation, consider a dataset of high-resolution human faces and corresponding genders.
A downstream model developer wants to develop an AI model that estimates skin health parameters and performs reliably irrespective of gender, skin tone, skin oiliness, and facial shape.
As a result, the model developer would like to train and evaluate their model on data that are representative of these four attributes.
The model developer works to curate a dataset of the three missing attributes given the faces, say by means of subject interviews and crowd-sourced data annotation.
The model developer now has two choices.
They can either partner with the original dataset owner to merge their new results with the original dataset, creating a new version of that dataset with updated dataset documentation, or create a new separate dataset with its own documentation that focuses on the new dataset while referencing the existing dataset documentation.
In both cases, the existing data documentation update or new data documentation would mainly focus on providing representation statistics of the three new attributes, information about the data-generating / collection process (including known annotation/annotator biases, etc.), and information about the intended and unintended usages of these data.
In the latter case, in which a new dataset and data documentation is created by the model developer, the model developer would also ideally be encouraged to contribute a ``Known Usage'' entry to the original data documentation.

\subsubsection{Dataset Status, Characteristics, Origin and Source}
The first section enables capturing the dataset’s maintenance status, core characteristics (e.g., data types, volume, number of instances), and intended AI use cases, including any associated systems. It also includes details on features, annotations, and target variables to support broad applicability. Provenance and sourcing are documented through descriptions of data origin, collection methods, platform, update frequency, and associated risks. This ensures transparency and reliability in understanding the dataset’s composition and lifecycle context.

\subsubsection{Data Pre-Processing; Versioning, Access, Retention and Deletion}
This section outlines key practices in data collection, pre-processing, and post-processing, including cleaning, transformation, feature engineering, dimensionality reduction, and augmentation. It documents annotation procedures and quality assurance to ensure labeling accuracy. Dataset licensing, distribution, and versioning are addressed to support transparency, traceability, and usability. Guidance on access, retention, and deletion is included to promote secure and accountable data management within research workflows. It is important to log this information to promote clarity, accountability, and secure data handling in research workflows.

\subsubsection{Data Risks and Security}
The cybersecurity section of the documentation template outlines robust protocols for safeguarding data. Key measures include documenting methods for secure storage through encryption, role-based access, and integrity monitoring; secure transfer and data masking; and secure processing using trusted execution environments, audit logging, and data minimization. These steps ensure data confidentiality, integrity, and compliance across its lifecycle.

\subsection{Model Documentation Template}
This template operationalizes the AI Act’s model documentation requirements by building on best practices for enabling structured reporting of intended and unintended uses, model architecture, training processes, hyperparameters, and evaluation metrics to enable deeper insight into model behavior~\cite{chudasama2023enhancing,kreuzberger2023machine,gong2023intended,tagliabue2021dag, arboretti2022design,kim2024explaining,sovrano2022metrics}. It builds on foundational work such as Model Cards~\cite{mitchell_model_2019}, Factsheets~\cite{arnold2019factsheets}, and other key frameworks~\cite{richards2020methodology,crisan2022interactive,golpayegani2024ai,hupont2024use}. 

As with the data documentation template, it is important to note that AI models, though designed for a particular AI system, are not necessarily associated with any one AI System. Thus, the model documentation template is designed to be filled without complete knowledge of all downstream AI Systems. Instead, the focus is on enabling model owners to make clear statements about the intended and unintended downstream usages of the model, potentially also with hypothetical examples of appropriate and inappropriate AI Systems.
As known usages become known, downstream developers that use the model are encouraged to contribute information to a ``Known Usages'' section.

As a generic example portraying the version-controlled, distributed approach to contributing to model documentation, consider again an AI model that estimates skin health parameters for dermatologists based on high-resolution face images (henceforth skin parameter model).
A downstream AI System developer wants to develop an AI System that provides Skincare advice based on the skin parameter model, as well as other models and logic, and wishes to fine-tune the existing skin parameter model to output a particular type of score.
The downstream system developer would then create the appropriate downstream AI system documentation based on the application documentation template, as well as model documentation for the new fine-tuned AI model.
In both cases, the new application and model documentation would refer to the existing model documentation, and the downstream AI System developer would also ideally be encouraged to contribute to a ``Known Usages'' section of the skin parameter model documentation. The following sections describe key sections of the model documentation in more detail.

\subsubsection{Model Overview, Intended Purpose, Architecture}
This section provides an overview of the model, including its architecture, key components, parameters, and selected training method, as well as duration and compute resources. It also documents input and output formats. The Model Purpose subsection defines the model’s intended and known applications, operational domains, and specific tasks, to support ethical and regulatory assessment. It includes known use cases and their associated risk levels under the AI Act, and it clearly identifies unsuitable applications to promote responsible and safe deployment.

\subsubsection{Model Validation}
The Model Validation section provides a structured approach for documenting model performance. It focuses on assessing predictions using a validation dataset, monitoring metrics like accuracy, F1 score, and RMSE, and tracking validation loss to detect overfitting that compromises generalization. It also includes benchmarking results, stress testing, and real-world performance assessments across diverse environments to ensure an overview of robustness. Key optimization strategies, such as hyperparameter tuning, regularization, and early stopping, can be documented to demonstrate efforts to enhance generalization and prevent overfitting. This part ensures a transparent, reproducible, and comprehensive validation process.

\subsubsection{Model Evaluation}
The Model Evaluation section in the template enables a comprehensive assessment of the model's performance using a variety of metrics and techniques. It begins with the computation of performance metrics on the test set, such as accuracy, precision, recall, and F1 score for classification tasks, or MSE, RMSE, and MAE for regression tasks. Moreover, the documentation of confusion matrices provides detailed insights into classification outcomes, while ROC curves and AUC scores offer a deeper evaluation of binary classifiers. Feature importance analysis enhances interpretability by highlighting key contributors to predictions. Robustness testing can be documented to assess performance on edge cases or adversarial examples, ensuring model reliability under challenging conditions. 

\subsubsection{Model Bias and Fairness}
The Bias Detection and Mitigation section in the template provides a structured framework to document efforts in identifying and addressing biases within a model.
This part was particularly challenging to draft as significant research has been conducted on the incompatibility of various fairness criteria in algorithmic decision making~\cite{loosley2023body,zehlike2025beyond}.
Hence, in this section, we build on the methods developed to interpolate between different fairness criteria \cite{zehlike2025beyond}, and understand what to document along the AI lifecycle, to understand where bias can enter ML models \cite{barocas2016big,friedman1996bias,corbett2023measure,barocas2023fairness}, for instance, through non-representative training data. 
This part provides a structured guide on bias detection methods across three stages: pre-processing (resampling, reweighting, relabeling), in-processing (transfer learning, constraint optimization, adversarial learning), and post-processing (calibration, thresholding). 
Results from bias testing can be recorded to assess the extent of disparities and develop mitigation strategies such as demographic parity and adversarial debiasing. Retraining methods, such as fairness regularization, recalibration, and output perturbation, can be documented to understand the model's performance. A Fairness Impact Statement concludes the documentation, outlining trade-offs made to ensure transparency and accountability.

\subsubsection{Model Transparency and Explainability}
This section documents the methods and tools used to make the model's decision-making process transparent and comprehensible, such as Shapley Values, LIME, or Counterfactual Explanations. This documentation ensures that stakeholders can assess the interpretability of the model and trust its decisions.

\subsection{Application Documentation Template}

This template focuses on documenting how AI models and other logic are integrated to form an application, including APIs, deployment environment, and user interaction.
This helps developers and end-users understand operational aspects of systems.
AI applications are complex systems that must be documented as they often include multiple components whose interaction significantly affects the overall performance and impact of the system~\cite{smart2024sociotechnical,lee2023collections,micheli2023landscape, madaio2020co}.
For this template, we build on the best practices promoting the documentation of systems in their entirety, with each component, such as factsheets~\cite{arnold2019factsheets,adkins2022method}, system cards \cite{procope2022system}.
Moreover, this section builds on the research analyzing the provision of evidence about the safety and security \cite{brundage2020toward}, fairness, accuracy, accountability, reliability, and privacy protection of AI systems \cite{brundage2020toward,li2023trustworthy,schoenherr2023designing,dobbe2022system}.

\subsubsection{AI Systems General Information, Risk Assessment, Functionality}
This section outlines the AI system’s purpose, target users, deployment context, and key performance goals, while addressing ethical considerations, prohibited uses, and AI Act–based risk classification. It describes system functionality, including capabilities, input requirements, usage scenarios, and limitations, with guidance on interpreting outputs. A high-level architecture overview covers core components and emphasizes integration with documentation, logging, and responsible contacts to support transparency and accountability.

\subsubsection{System Deployment}
The Deployment section provides a concise framework for documenting the infrastructure, integration, and operational requirements of the AI system. It outlines the deployment environment and the integration with external systems is detailed through dependencies, data flow diagrams, and error-handling mechanisms. The section also includes a deployment plan covering infrastructure, steps, and security compliance, alongside a monitoring framework to track performance. Finally, it provides user documentation to support operators and end-users in effectively using the system.

\subsubsection{Lifecycle Management}
The Lifecycle Management section documents procedures for monitoring performance to comply with the post-market monitoring requirement of the AI Act, ethical compliance, and versioning throughout the system's lifecycle. It outlines metrics for application performance (e.g., response time, error rate), model accuracy, and infrastructure usage. Key activities that must be described include real-world monitoring, addressing drifts or failures, and periodic model updates. Monitoring logs, incident reports, retraining logs, and audit trails can be inserted here to ensure transparency, compliance, and continuous improvement.

\subsubsection{Risk and Incident Management}
The Risk Management can be documented according to the Article 9 requirements, by outlining measures to ensure safe and ethical operation along frameworks like ISO 31000 or NIST \cite{NIST_SP_800-37r2}. It enables documenting identified risks and potential harmful outcomes (e.g., bias, privacy breaches) and assessing their likelihood and severity. Preventive measures, such as data validation and bias mitigation, can be documented alongside protective measures, including contingency plans to minimize impact. It covers the documentation of all possible incidents, such as infrastructure challenges alongside integration problems. Strategies for ensuring data quality, handling model issues like drift, and addressing safety or security risks can be delineated. 

\subsubsection{Testing and Validation (Accuracy, Robustness, Cybersecurity)}
The Testing and Validation section ensures the AI system’s reliability is documented through rigorous evaluation of accuracy, robustness, and cybersecurity. It outlines performance metrics (e.g., accuracy, F1 score) and validation results against benchmarks, with measures about the data curation, algorithm optimization, and real-time feedback to maintain accuracy. Robustness can be documented through the deployment of adversarial training, stress testing, and fail-safes to handle edge cases and uncertainty. Cybersecurity documentation focuses on threat modelling, secure development, and post-deployment monitoring, with detailed documentation to ensure compliance and accountability.

\subsubsection{Human Oversight}
The Human Oversight section follows the requirements of Articles 13 and 14 of the AI Act \cite{EU_AI_Act_2024}. Details mechanisms for integrating human judgment into the AI system, such as human-in-the-loop decision-making and override options for emergencies, can be documented. It includes information about user training and space for guidelines to ensure safe operation, along with the documentation of clear statements of the system's limitations and potential weaknesses to promote responsible use.

\section{Validation}
The templates are tested through their implementation and rendering on real-world example scenarios. 
Such practical example implementations augment the templates to provide further guidance to documentation developers. In the data documentation example, a skin tone dataset used in an E-Commerce AI system is documented to highlight potential biases and support fairness evaluations \cite{loosley2023body}. This use case validates the utility of the TechOps approach as it provides an overview of all the necessary information to help downstream stakeholders understand the dataset, providing information about the distribution of skin tones in the customer images dataset. In fact, the documentation readers can easily find out that the dataset is made up primarily of light skin tone users and would potentially underperform for users with underrepresented skin tones.
In the model documentation example, the performance and limitations of the ALiSNet segmentation model are documented to guide downstream developers in identifying risks, particularly related to body shape bias. Both examples show how the templates help developers meet regulatory requirements and conduct more informed system evaluations \cite{seifoddini2023alisnet}. 

\subsection{Data Documentation Example}
The context of this example is an E-Commerce company developing an AI System to improve size and fit recommendations based on customer images to estimate body measurements \cite{loosley2023body}.
In the real example, the E-Commerce company conducted a fairness evaluation to ensure the system did not systematically underperform for customers of certain genders, body shapes, or skin tones.
A Skin Tones Dataset based on human annotations of skin tones of real customer images was curated to carry out this fairness evaluation and is the subject of this data documentation example.

The skin tones dataset documentation provides an overview of all the necessary information to help downstream stakeholders understand the dataset. It provides information about the distribution of skin tones in the customer images dataset,
the potential biases arising from the data-generating process (in this case, human annotation), and the intended purpose and usage of the dataset.
Documentation readers can, for example, easily find that the dataset is made up primarily of light skin tone users.
This documentation can be used to help developers of downstream AI Systems identify and assess risks, such as the system potentially underperforming for users with underrepresented skin tones.
Ultimately, for providers of high-risk AI Systems based on components trained or tested with the skin tones dataset, the skin tones data documentation can be used to, in part, comply with the technical documentation requirements set forth in Annex IV in order for them to pass a conformity assessment.

\subsection{Model Documentation Example}
The context of this example is again an E-Commerce company developing an AI System to improve size and fit recommendations based on customer images to estimate body measurements.
One key component of the AI System is a neural network-based image segmentation model called ALiSNet that outputs silhouette images of customers based on front and side photos the customer takes with their mobile phone \cite{seifoddini2023alisnet}.

The ALiSNet model documentation provides an overview of all the necessary information to help downstream stakeholders understand the model, including its intended use, how well it performs overall, under what circumstances it is prone to systematic underperformance, known caveats, and ethical considerations.
Documentation readers can, for example, easily find that the model performs well regardless of gender and skin tone, but there is a statistically significant detectable difference in how the model performs with respect to body shape.
Despite this, on an absolute scale, the underperformance for some body shapes is not likely severe because the absolute average length scale of error of silhouettes obtained by the model on images of customers with the weakest performing body shapes was still below the typical body measurement error of a trained human taking measurements.
Thus, this documentation helps downstream AI System developers, like those developing an AI System for Size and Fit recommendations, identify and assess potential risks, such as the system potentially underperforming for users of certain body shapes.
Even though the underperformance on specific body shapes is minimal from the AI Model itself, even small errors can be amplified as they propagate through the various components making up the AI System.
Thus, the AI System developer is informed by this model documentation that they should probably also test their AI System end-to-end to ensure that the system overall does not underperform for certain body shapes.

This model documentation not only guides the downstream AI System on how they may want to design a fairness evaluation, but also documents crucial data sources the AI System developer can use to perform the fairness evaluation.
Thus, the AI System developer does not have to reinvent the wheel.

Ultimately, for providers of high-risk AI systems that use ALiSNet as one of their components, the model documentation can be used to, in part, comply with the technical documentation requirements in Annex IV in order to pass a conformity assessment.

\begin{figure}
	\vspace{0mm}
	\centering
	\includegraphics[width=0.5\textwidth]{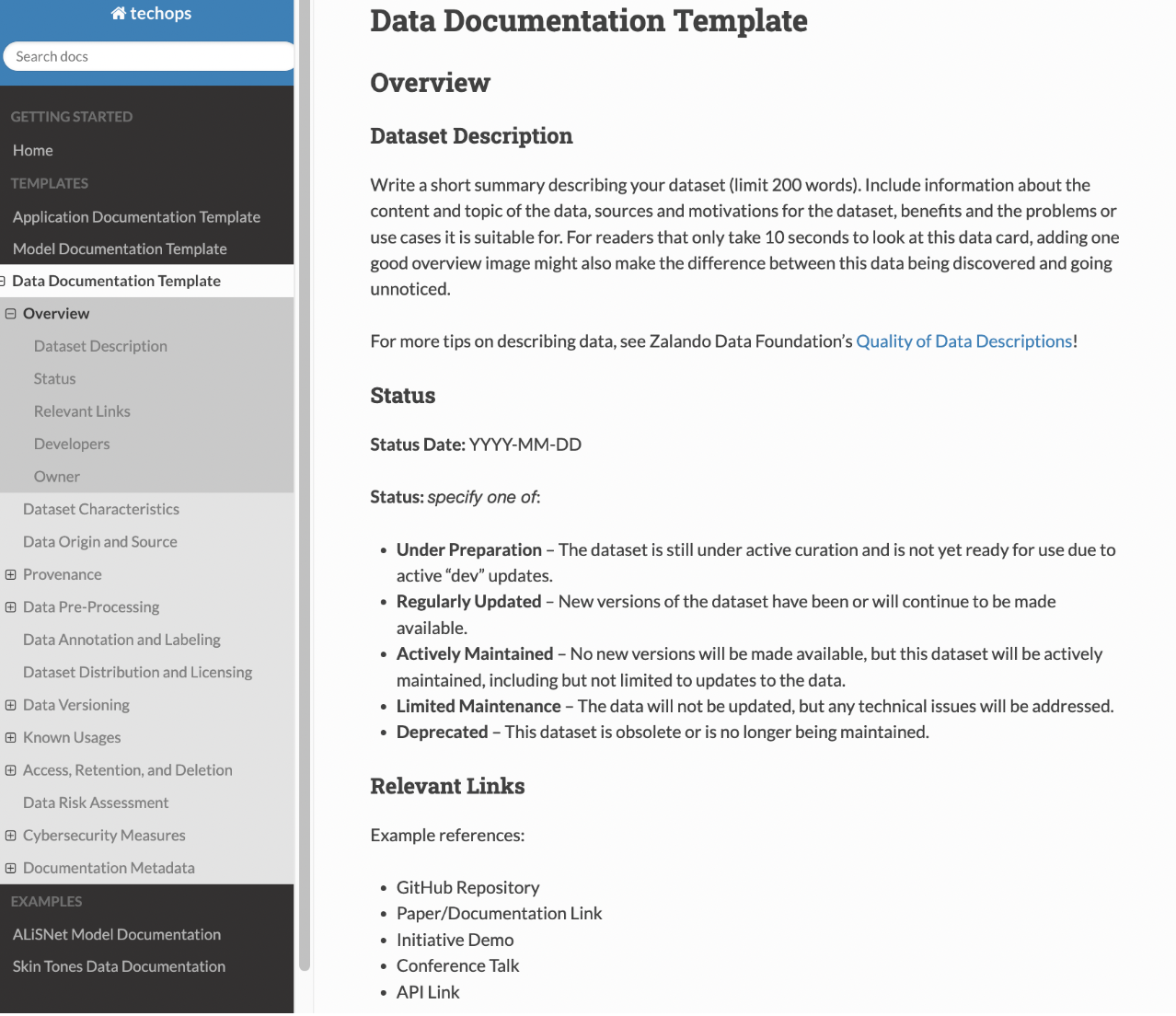}
        \caption{\textbf{Template and example rendering snippet.} (See \url{https://aloosley.github.io/techops/})}
	\vspace{-3mm}
	\label{fig:rendering}
\end{figure}

\subsection{Application Documentation Example}
The application documentation example is applied to a high-risk AI system developed to enable construction site safety using real-time video analytics and sensor data. Deployed on-site with edge devices and cloud integration, it identifies unsafe behaviors (like lack of PPE or entering danger zones) and hazardous conditions to trigger alerts, reduce accidents, and support regulatory compliance. The documentation process enables gaining insights into whether the application is designed for site supervisors and safety managers or whether the system aligns with GDPR and the EU AI Act, with human oversight built in for alert validation. This process provided critical visibility into its functionality, data flows, model limitations, and associated risks, enabling better governance, regulatory alignment, and continuous system monitoring and improvement.

\subsection{Rendering of Templates and Examples}
To ensure that both documentation developers and users can optimally view complex documentation without getting lost in the details, we provide an mkdocs based open source rendering so that one can always maintain a documentation overview while deep diving into specific sections of interest (\url{https://github.com/aloosley/techops}). For convenience, we have also rendered all documentation templates and examples online at \url{https://aloosley.github.io/techops/} (see Fig.~\ref{fig:rendering}).

\section{Evaluation and Feedback}
This section offers a thematic analysis of the qualitative insights gained from conducting semi-structured interviews with users and practitioners along the AI lifecycle \cite{clarke2017thematic,terry2017thematic,braun2022conceptual,squires2023thematic}. The practitioners were recruited through theoretical sampling, which required participants in the study to work along the AI lifecycle as either an ML engineer, a data scientist, a legal expert, or an auditor. This methodology enabled us to collect feedback from the relevant stakeholders around the AI lifecycle and make sure that the templates enable all information relevant for different stakeholders (see Tab.~\ref{tab:interview}). 
Due to the novelty of the research topic in best practices for AI governance, the “snowball” technique was applied to expand the network through the connections of the interview candidates \cite {naderifar2017snowball}.
We used an inductive thematic analysis to organically identify and analyse patterns in the stakeholder feedback data to capture all potential improvement suggestions that would enable us to increase the templates' implementability \cite{braun2024thematic, braun2022conceptual}.
\begin{table}
  \caption{User Study Participants}
  \label{tab:interview}
  \centering
  \small
  \begin{tabular}{llll}

    \bf ID & \bf Role & \bf Organization & \bf Background \\
    U1  & Researcher           & Industry   & Mathematics       \\
    U2  & Senior Researcher    & Robotics   & Informatics  \\
    U3  & Researcher           & Industry   & Informatics  \\
    U4  & Senior Researcher    & Industry   & Informatics  \\
    U5  & Senior Researcher    & Industry   & Informatics  \\
    U6  & Manager              & Industry   & Informatics  \\
    U7  & Researcher           & Industry   & Informatics  \\
    U8  & Senior Developer     & Industry   & Informatics  \\
    U9  & Senior Developer     & Industry   & Informatics       \\
    U10 & Team Lead            & Industry   & Informatics       \\
    U11 & Senior Data Scientist& Finance    & Data Science      \\
    U12 & Project Manager      & Banking    & Economics         \\
    U13 & Researcher           & Tech       & Mathematics       \\
    U14 & Researcher           & Academia   & Law               \\
    U15 & Researcher           & Academia   & Law               \\
  \end{tabular}
\end{table}

  \subsection{Theme 1: Avoiding Duplication of Information}
One of the main issues that emerged is the duplication of information that can occur when navigating separate templates of intertwined components, which might lead to unsustainable maintenance issues (User 1, 2, 3, 4, 5, 6, 8, 9). In order to avoid duplication of information, we have color-coded in the application template the information that must be retrieved from the data and model documentation in order to enable maintenance in the long term, while keeping a clear oversight on the data and model selection practices that impact the overall quality and impact of the application. Moreover, some users found this approach helpful in documenting beyond compliance needs for business and research purposes (User 5, 13, 14). For instance, the templates could be useful to fill the documentation requirements ``when publishing research on AI'' (U5) or when ``discussing with clients the AI system provided'' (U13). 

\subsection{Theme 2: Implementability Issues}
Users were concerned with the comprehensiveness of the templates covering all tasks along the AI lifecycle, which would make the templates too overwhelming for practitioners focusing on a specific task (User 1, 2, 3, 5, 7, 8, 9). Hence, not just a task related but use-case related automation would be necessary for customization (User 1, 5, 7, 8, 9). 
Practitioners also expressed their concerns about making sure that the templates can be easily implemented within the organizational workflows to optimize their usability and maintenance (User 1, 2, 3). In practice, developers would mainly be able to log relevant information if it is automatically generated whilst coding and send this information directly to a report. Hence, this framework can help developers understand which decisions must be logged and why, and can serve as a guiding framework that collects all relevant information to be documented. 
A solution could be to adopt methods such as the automatic versioning of Git, like GitHub Action, where practitioners can check the git repository to see who worked on which part of the templates. Hence, we made sure that automating our documentation is fully achievable. 
Compared to existing templates, certain users (User 1, 5, 7, 8, 9) perceived Techops to be easier to follow from a technical perspective, which enables them to better document their code, as other templates, such as datasheets, are more abstract and therefore more challenging to fill. Moreover, some users (U12, U13, U14, U15) found the templates approach helpful in navigating interdisciplinary teams' barriers to understanding the translation of abstract concepts into actionable processes along the lifecycle. 

\subsection{Theme 3: Guidance for documentation and understanding implications}
A theme that emerged is that many practitioners, especially those without a legal background, are unaware of the documentation requirements of the AI Act (User 1--13). This clearly delineates the need to implement better governance strategies across the companies to clarify what must be documented. However, this is currently challenging due to the lack of standards and open-source solutions. A clear difference we could observe was the awareness of adopting such practices early on between the users in startups and those in highly regulated industries such as banking and finance (U11, U12). This finding suggests that SMEs lack the resources currently on the market to implement such solutions and understand the implications of good governance practices early on. 
Certain users needed more information and examples to understand the implications of deploying certain systems,as well as the necessity to document certain processes along the lifecycle (User 5, 7). For instance, certain developers did not understand the necessity for documenting specific details such as the hyperparameters (User 3, 5). We had to explain that in highly regulated fields like finance, for instance, inspecting hyperparameters enables us to see how potentially biased a model is.  Hence, such inspection is necessary to avoid overfitting of sensitive attributes and making models simpler and less likely to memorize biased patterns. Another User (3) was concerned with the initial sections defining in and out of scope use-cases, as it is hard to predict for developers inappropriate usage.  However, this process is fundamental to defining liability and delineating the safe extent of deployment of certain systems by outlining what they were optimised for. We included examples to guide users in delineating the safe extent of deployment. Moreover, certain users (User 4, 5) needed more guidance on the distinction between documenting necessary information for governance purposes, whilst avoiding the documentation of important intellectual property information. 
We clarified how to keep the balance in the templates, such as how model type and architecture can be described without disclosing business-critical information,such as fine-tuning strategies or optimization methods. 

In all our tests and feedback sessions, a significant issue raised by the users has been the absence of open-source templates and standards targeting the AI Act’s documentation requirements throughout the entire lifecycle. Many companies openly struggle to meet these requirements and remain unsure about documentation standards. Users expressed concern over the perceived gaps in current AI compliance solutions, which often lack guarantees and remain unaffordable for SMEs.
We therefore stress the importance of guidance from the open-source community. The current best practices in documentation such as Datasheets for Datasets \cite{gebru_datasheets_2021} and Model Cards \cite{mitchell_model_2019} or Factsheets \cite{arnold2019factsheets} or Data Cards\cite{pushkarna2022data} provide a useful starting point by suggesting important questions that should be answered by all the actors involved in the AI lifecycle. Moreover, existing frameworks that propose documentation templates for AI Act compliance such as AI Cards \cite{golpayegani2024ai} and Use Case Cards \cite{hupont2024use} do not fully capture the intricacies of AI system components or the technical practices applied across the lifecycle, including design, development, deployment, and post-market monitoring. This limitation makes it difficult for large organizations with diverse stakeholders to develop a comprehensive understanding of these systems.
Existing best practices in AI documentation do not capture the entire AI lifecycle while aligning with the specific legal obligations related to risk management, data governance, human oversight, robustness, and accuracy, as well as a framework for continuous monitoring and lifecycle updates.
Moreover, existing practices lack the legally required level of detail to document system interactions, software versions, deployment methods, and human oversight measures.
 Compared to the existing open-source templates, users found TechOps to help bridge the gap between abstract legal requirements and actual documenting of practices along the AI lifecycle. For instance, compared to questionnaire-based approaches such as datasheets, Techops was found to be better at enabling this overview for different stakeholders and guiding technical practitioners to understand the implications of their choices.  The templates provided in this paper capture decision-making processes across the AI lifecycle, from design to post-market monitoring. They do so by tracking performance metrics over time, as everything is under version control so it is easy to identify causes of variation, and benchmark across models.
Finally, they improve auditability by providing a central repository of all relevant information, simplifying the audit process for both auditors and auditees. By integrating these structured practices, the proposed documentation templates not only facilitate compliance with the AI Act but also enhance fairness, trust, transparency, and accountability in AI system development.

 Our findings show that practitioners still need help applying documentation solutions consistently throughout the AI system lifecycle, as this is not yet a common practice. There is a significant gap between legal and technical experts in understanding why certain elements must be documented and what their legal and technical consequences are. This combined expertise is rare and typically found in auditors. To address this gap, the templates should be reviewed in multidisciplinary meetings at each stage of the lifecycle to ensure effective auditing and oversight.

TechOps can guide stakeholders across the entire AI lifecycle, serving not only as a tool for documenting compliance with the AI Act but also as a framework for fostering the development of safe, robust, fair, and reliable AI systems. By systematically bridging the gap between legal and technical literacy, this approach mitigates the risk of ethics washing, ensuring that the completion of documentation templates involves substantive reflection on both the functional characteristics of the system and its legal implications.
Our users said that they could also benefit from using TechOps to submit documentation when publishing AI research or open-sourcing their AI systems, models, and datasets. Therefore,  we hope that TechOps provides a standardised and adaptable framework that contributes to emerging best practices. We hope it serves as a building block toward a more consistent and thoughtful approach to AI development---one that not only meets regulatory demands but also fosters trust, accountability, and collaboration within the broader research and developer communities.

\subsection{Limitations}
AI systems are currently deployed in many domains, which means that these templates are not to be seen as a  one-size-fits-all solution but as a contribution to enable relevant stakeholders to understand what to document throughout the AI lifecycle. The aim was to provide guidance through the translation of the legal requirements; however, it is unlikely that these templates can completely capture all components and practices across the different vertical domains in which AI is deployed. 
Moreover, we are aware of the complexity vs.\ applicability dilemma that such thorough documentation is time-consuming and might not be sustainable and maintainable. 
However, we believe that without thorough guidance and an overview of the components and processes of the AI lifecycle, efficient governance measures cannot be implemented. We thus encourage organizations and relevant stakeholders to adapt these templates to their own context of deployment.

Documenting datasets, models, and applications to comply with the AI Act in the industry still remains a challenge due to the lack of resources and solutions, constant innovation, iterative design and engineering approaches, lack of reliable testing resources, and many other reasons that make documentation a challenge. Here, we provide open-source templates that can be automated, addressing a clear gap in governance and oversight practices to help companies navigate legal uncertainties around AI Act documentation.

\section{Conclusion}
We propose TechOps: templates and examples for documenting data, models, and applications to provide sufficient documentation for AI Act certification. TechOps addresses the gaps between the existing
documentation templates by enabling a holistic approach to documenting all components and processes across the lifecycle while meeting the technical documentation requirements of the AI Act. 
We believe this work to be a key input in the ongoing efforts to operationalize the AI Act requirements. Beyond regulatory alignment, the templates serve as practical tools for governing AI systems by helping responsible stakeholders assess quality metrics, identify risks and biases, and prevent the accumulation of technical debt throughout the system lifecycle.

\section{Acknowledgments}
We would like to thank Pak-Hang Wong, Rocco Maresca, Scott Small, Daniela Kato, Sobia Naz, Silke Claus, Hosna Sattar, Amrollah Seifoddini, Mathias Deschamps, Ola Wahab, Stanimir Dragiev, and all the other applied scientists, engineers, lawyers, and business leaders that provided valuable feedback. Their insights were instrumental in shaping this work to meet the diverse needs of developers and the many types of users and stakeholders of AI/ML systems. We would like to thank Chiara Gei for her significant contributions on model cards in etami, which played a key role in building this work, and to the etami consortium members for their foundational efforts in translating ethical AI principles into actionable practices. 

\section{Appendices}

\section{Templates Iteration}
\subsection{Version 1}
In the first iteration of the templates with the users and practitioners feedback was the lack of guidance of understanding the legal and ethical implications of documenting certain parts such as the hyperparameter tuning. For instance, documenting hyperparameter tuning is essential in sensitive domains because these parameters directly influence model fairness, interpretability. For example, in loan approval prediction use-cases if systems are tuned only to maximize AUC without fairness metrics, the model may suppress approval rates for marginalized groups whose profiles resemble historically rejected applicants.

On the other hand, we needed to add these examples to guide non technical stakeholder such as governance, legal experts and auditors to understand the implications of certain practices along the lifecycle.
We realised that to solve this issue we needed to insert descriptive sections in the templates to guide practitioners with real-work examples to understand why documenting certain information is crucial for safety and accountability later on. 
\subsection{Version 2}
The primary concerns we addressed in the second iteration were the lack of delineation of which information was necessary for documentation whilst preserving intellectual property as well as which information is necessary to define liability. We realised that we needed to provide guidance in the templates on how to describe certain methods. For instance U5, U8 and U9 wanted to understand which information is necessary for documentation as they might not be able to ``provide details on practices they were not directly responsible for'', for instance when you post-process data that you acquired from third parties.  Therefore, in the descriptive parts of the templates we delineated the AI Act articles requirements as well as some examples of how to document certain processes without the need to share intellectual property information.

\subsection{Version 3}
The third iteration had to solve the language barrier of the interdisciplinary stakeholders as well as delineate who is responsible for documenting which processes. We therefore inserted info sections on the sections of the templates explaining what the law requires and we then illustrate the technical processes that correspond to those abstract legal requirements. This was necessary to enable multidisciplinary teams along the AI lifecycle to be able to communicate more effectively with each other and create a bridge to translate abstract legal requirements into practices that must be documented along the lifecycle

\section{Implementation Guidance}

We hope this document will provide a solid basis to implement documentation practices across different domains. TechOps should serve as a guidance on the essential practices that must be documented along the lifecycle pipeline in order to enable compliance with the AI Act requirements. 
It was important to use technical language to guide documentation as ultimately all the governance and oversight mechanisms must start with understanding the technical measures adopted along the lifecycle and make inferences about what certain scores of accuracy and precision, for instance, mean in practice and therefore to quantify potential risks. 
Therefore, to understand how risks emerge and what decisions might have caused them it is important to have a detailed technical overview of the important processes taken along the lifecycle. 

\subsection{Templates}

TechOps consists of three separate templates for  documenting AI systems to provide proof of compliance with the AI Act.
The documentation is split into three levels:

\begin{itemize}
\item AI System documentation
\item Model documentation
\item Data documentation
\end{itemize}

The separate templates allow the owners of data, models, and AI systems to each maintain ownership of their own level of documentation.  
Thus, model and dataset owners who may or may not have curated their models and datasets with a specific AI System in mind, may still create documentation contributions that the AI System documentation can reference.

These templates are meant to guide responsible stakeholders in documenting AI systems across various domains. 
Unlike existing lengthy and abstract questionnaires, these templates offer clear guidance for the documentation of the relevant processes across the AI lifecycle, translating complex requirements such as fairness and data governance into actionable metrics and measurable criteria that can be implemented and tracked. 
This process ensures that the abstract legal requirements of the AI Act are operationalized into concrete actions, making them manageable and measurable.
Following the TechOps approach also provides stakeholders comprehensive oversight on the data, model and application lifecycle. These templates track the system’s status over the entire AI lifecycle, ensuring traceability, reproducibility, and compliance with the AI Act. Clear documentation also promotes discoverability, collaboration, and risk assessment.

\subsection{How To Use the Templates}

The implementation of the TechOps framework follows a structured process designed to ensure that the resulting documentation is both comprehensive and contextually relevant. The process begins with the identification of the appropriate documentation scope. TechOps provides three templates, corresponding to the \emph{AI Application}, \emph{Models}, and \emph{Data} levels.

The approach is to choose which of the templates need to be
filled. For example, if you are responsible for an AI System
and have trained your own models based on custom datasets
that have not been documented, you may need to work at all
three documentation levels (AI System, model, and data).
We, the TechOps developers, advise that documentation
authors use the templates as best suites their needs given
documentation standards that already exist in your context.
One example of how to use the templates would be, first, to create a GitHub Repo for your documentation. Then, you can access the template(s) markdown file you wish to implement, see [our GitHub Repo](\url{https://github.com/aloosley/techops}) and implement them with the help of the template comments and examples. Rendering is suggested to enable a better overview. 
We provide an mkdocs based open source rendering so that one can always maintain a documentation overview while deep diving into specific sections of interest (\url{https://github.com/aloosley/techops}). For convenience, we have also rendered all documentation templates and examples online at \url{https://aloosley.github.io/techops/}.

\section{Motivation}

The AI Act mandates that technical documentation include key information such as a general description of the system, detailed accounts of its components, and a record of the processes conducted during its development. However, a significant challenge in the market is the absence of standardized documentation methods for ensuring compliance. This lack of uniformity often results in stakeholders independently creating documentation, a task frequently undertaken by legal teams whose language and focus differ from those of developers and other technical stakeholders. To bridge this gap, the proposed template is designed to accommodate the language and needs of all stakeholders involved in the AI lifecycle, ensuring that practices are clearly described and responsibilities are effectively communicated.

The proposed templates also address the requirements of companies that not only develop AI systems in-house but also acquire systems for deployment. By providing a comprehensive yet adaptable structure, the templates simplify the compliance process, enabling organizations to demonstrate adherence to authorities or acquirers efficiently. Covering all phases of the machine learning lifecycle, the templates are aligned with legal requirements while remaining practical for implementation.

To be effective, the templates integrate the perspectives of diverse stakeholders, including developers, managers, auditors, and legal experts. This approach bridges the abstract legal requirements of the AI Act with the technical practices necessary for ensuring risk assessment, data quality, and accuracy. Furthermore, the templates are designed as a domain-agnostic blueprint, serving as a foundation for documenting AI systems across various fields. At the same time, conciseness is maintained to avoid overwhelming stakeholders with excessive content. Unlike existing lengthy and abstract questionnaires, these templates offer clear guidance for the documentation of the relevant processes across the AI lifecycle, translating complex concepts such as fairness and data governance into actionable metrics and measurable criteria that can be implemented and tracked. This process ensures that the abstract legal requirements of the AI Act are operationalized into concrete actions, making them manageable and measurable.

In addition to ensuring compliance, structured documentation templates provide numerous benefits for AI development. By centralizing critical information such as training data, features, hyperparameters, and model types, they facilitate reproducibility and explainability. Transparency is enhanced through the documentation of the model’s development context, intended uses, and limitations, ensuring stakeholders understand how the system operates and its constraints. These templates also reduce technical debt by fostering early alignment among stakeholders, clarifying goals and limitations, and enabling timely adjustments, such as refining scope or collecting additional data. Standardization, supported by existing literature, ensures that documentation processes are flexible yet consistent, enabling partial automation where feasible. Accountability is strengthened as documentation of the key components of AI systems is recorded from all relevant stakeholders in the AI lifecycle. Queryability allows stakeholders to compare existing solutions, benchmark performance, and streamline the development of new systems for similar use cases. The templates also support monitoring by tracking performance metrics over time, identifying causes of variation, and enabling benchmarking across models. Finally, they improve auditability by providing a central repository of all relevant information, simplifying the audit process for both auditors and auditees.

By integrating these structured practices, the proposed documentation templates not only facilitate compliance with the AI Act but also enhance fairness, trust, accountability, and efficiency in AI system development.

\bibliography{aaai25.bib}

\begin{thebibliography}{92}
\providecommand{\natexlab}[1]{#1}

\bibitem[{Adkins et~al.(2022)Adkins, Alsallakh, Cheema, Kokhlikyan, McReynolds,
  Mishra, Procope, Sawruk, Wang, and Zvyagina}]{adkins2022method}
Adkins, D.; Alsallakh, B.; Cheema, A.; Kokhlikyan, N.; McReynolds, E.; Mishra,
  P.; Procope, C.; Sawruk, J.; Wang, E.; and Zvyagina, P. 2022.
\newblock Method cards for prescriptive machine-learning transparency.
\newblock In \emph{Proceedings of the 1st International Conference on AI
  Engineering: Software Engineering for AI}, 90--100.

\bibitem[{Afzal et~al.(2021)Afzal, Rajmohan, Kesarwani, Mehta, and
  Patel}]{afzal2021data}
Afzal, S.; Rajmohan, C.; Kesarwani, M.; Mehta, S.; and Patel, H. 2021.
\newblock Data readiness report.
\newblock In \emph{2021 IEEE international conference on smart data services
  (SMDS)}, 42--51. IEEE.

\bibitem[{Alder et~al.(2024)Alder, Ebert, Herbrich, and Hacker}]{alder2024ai}
Alder, N.; Ebert, K.; Herbrich, R.; and Hacker, P. 2024.
\newblock Ai, climate, and transparency: Operationalizing and improving the ai
  act.
\newblock \emph{arXiv preprint arXiv:2409.07471}.

\bibitem[{Arboretti et~al.(2022)Arboretti, Ceccato, Pegoraro, and
  Salmaso}]{arboretti2022design}
Arboretti, R.; Ceccato, R.; Pegoraro, L.; and Salmaso, L. 2022.
\newblock Design choice and machine learning model performances.
\newblock \emph{Quality and Reliability Engineering International}, 38(7):
  3357--3378.

\bibitem[{Arnold et~al.(2019)Arnold, Bellamy, Hind, Houde, Mehta,
  Mojsilovi{\'c}, Nair, Ramamurthy, Olteanu, Piorkowski
  et~al.}]{arnold2019factsheets}
Arnold, M.; Bellamy, R.~K.; Hind, M.; Houde, S.; Mehta, S.; Mojsilovi{\'c}, A.;
  Nair, R.; Ramamurthy, K.~N.; Olteanu, A.; Piorkowski, D.; et~al. 2019.
\newblock FactSheets: Increasing trust in AI services through supplier's
  declarations of conformity.
\newblock \emph{IBM Journal of Research and Development}, 63(4/5): 6--1.

\bibitem[{Arnold et~al.(2024)Arnold, Yesilbas, Gr{\"o}bner, Riedelbauch, Horn,
  and Weinzierl}]{arnold2024documentation}
Arnold, S.; Yesilbas, D.; Gr{\"o}bner, R.; Riedelbauch, D.; Horn, M.; and
  Weinzierl, S. 2024.
\newblock Documentation Practices of Artificial Intelligence.
\newblock \emph{arXiv preprint arXiv:2406.18620}.

\bibitem[{Assembly(2024)}]{ColoradoAIAct2024}
Assembly, C.~G. 2024.
\newblock Senate Bill 24-205: Consumer Protections for Artificial Intelligence.
\newblock Accessed: 20 January 2025.

\bibitem[{Barocas, Hardt, and Narayanan(2023)}]{barocas2023fairness}
Barocas, S.; Hardt, M.; and Narayanan, A. 2023.
\newblock \emph{Fairness and machine learning: Limitations and opportunities}.
\newblock MIT press.

\bibitem[{Barocas and Selbst(2016)}]{barocas2016big}
Barocas, S.; and Selbst, A.~D. 2016.
\newblock Big data's disparate impact.
\newblock \emph{Calif. L. Rev.}, 104: 671.

\bibitem[{Bender and Friedman(2018)}]{bender2018data}
Bender, E.~M.; and Friedman, B. 2018.
\newblock Data statements for natural language processing: Toward mitigating
  system bias and enabling better science.
\newblock \emph{Transactions of the Association for Computational Linguistics},
  6: 587--604.

\bibitem[{Bhardwaj et~al.(2024{\natexlab{a}})Bhardwaj, Gujral, Wu, Zogheib,
  Maharaj, and Becker}]{bhardwaj2024machine}
Bhardwaj, E.; Gujral, H.; Wu, S.; Zogheib, C.; Maharaj, T.; and Becker, C.
  2024{\natexlab{a}}.
\newblock Machine learning data practices through a data curation lens: An
  evaluation framework.
\newblock In \emph{The 2024 ACM Conference on Fairness, Accountability, and
  Transparency}, 1055--1067.

\bibitem[{Bhardwaj et~al.(2024{\natexlab{b}})Bhardwaj, Gujral, Wu, Zogheib,
  Maharaj, and Becker}]{bhardwaj2024state}
Bhardwaj, E.; Gujral, H.; Wu, S.; Zogheib, C.; Maharaj, T.; and Becker, C.
  2024{\natexlab{b}}.
\newblock The State of Data Curation at NeurIPS: An Assessment of Dataset
  Development Practices in the Datasets and Benchmarks Track.
\newblock \emph{arXiv preprint arXiv:2410.22473}.

\bibitem[{Billeter et~al.(2024)Billeter, Denzel, Chavarriaga, Forster,
  Schilling, Brunner, Frischknecht-Gruber, Reif, and Weng}]{billeter2024mlops}
Billeter, Y.; Denzel, P.; Chavarriaga, R.; Forster, O.; Schilling, F.-P.;
  Brunner, S.; Frischknecht-Gruber, C.; Reif, M.; and Weng, J. 2024.
\newblock MLOps as enabler of trustworthy AI.
\newblock In \emph{2024 11th IEEE Swiss Conference on Data Science (SDS)},
  37--40. IEEE.

\bibitem[{Birkstedt et~al.(2023)Birkstedt, Minkkinen, Tandon, and
  M{\"a}ntym{\"a}ki}]{birkstedt2023ai}
Birkstedt, T.; Minkkinen, M.; Tandon, A.; and M{\"a}ntym{\"a}ki, M. 2023.
\newblock AI governance: themes, knowledge gaps and future agendas.
\newblock \emph{Internet Research}, 33(7): 133--167.

\bibitem[{Braun and Clarke(2022)}]{braun2022conceptual}
Braun, V.; and Clarke, V. 2022.
\newblock Conceptual and design thinking for thematic analysis.
\newblock \emph{Qualitative psychology}, 9(1): 3.

\bibitem[{Braun and Clarke(2024)}]{braun2024thematic}
Braun, V.; and Clarke, V. 2024.
\newblock Thematic analysis.
\newblock In \emph{Encyclopedia of quality of life and well-being research},
  7187--7193. Springer.

\bibitem[{Brundage(2019)}]{brundage2019responsible}
Brundage, M. 2019.
\newblock Responsible governance of artificial intelligence: An assessment,
  theoretical framework, and exploration.
\newblock Technical report, Arizona State University.

\bibitem[{Brundage et~al.(2020)Brundage, Avin, Wang, Belfield, Krueger,
  Hadfield, Khlaaf, Yang, Toner, Fong et~al.}]{brundage2020toward}
Brundage, M.; Avin, S.; Wang, J.; Belfield, H.; Krueger, G.; Hadfield, G.;
  Khlaaf, H.; Yang, J.; Toner, H.; Fong, R.; et~al. 2020.
\newblock Toward trustworthy AI development: mechanisms for supporting
  verifiable claims.
\newblock \emph{arXiv preprint arXiv:2004.07213}.

\bibitem[{Castelijns, Maas, and Vanschoren(2020)}]{castelijns2020abc}
Castelijns, L.~A.; Maas, Y.; and Vanschoren, J. 2020.
\newblock The abc of data: A classifying framework for data readiness.
\newblock In \emph{Machine Learning and Knowledge Discovery in Databases:
  International Workshops of ECML PKDD 2019, W{\"u}rzburg, Germany, September
  16--20, 2019, Proceedings, Part I}, 3--16. Springer.

\bibitem[{Chmielinski et~al.(2024)Chmielinski, Newman, Kranzinger, Hind,
  Vaughan, Mitchell, Stoyanovich, McMillan-Major, McReynolds, Esfahany
  et~al.}]{chmielinski2024clear}
Chmielinski, K.; Newman, S.; Kranzinger, C.~N.; Hind, M.; Vaughan, J.~W.;
  Mitchell, M.; Stoyanovich, J.; McMillan-Major, A.; McReynolds, E.; Esfahany,
  K.; et~al. 2024.
\newblock The CLeAR Documentation Framework for AI Transparency.

\bibitem[{Chmielinski et~al.(2022)Chmielinski, Newman, Taylor, Joseph, Thomas,
  Yurkofsky, and Qiu}]{chmielinski2022dataset}
Chmielinski, K.~S.; Newman, S.; Taylor, M.; Joseph, J.; Thomas, K.; Yurkofsky,
  J.; and Qiu, Y.~C. 2022.
\newblock The dataset nutrition label (2nd gen): Leveraging context to mitigate
  harms in artificial intelligence.
\newblock \emph{arXiv preprint arXiv:2201.03954}.

\bibitem[{Chudasama et~al.(2023)Chudasama, Purohit, Rohde, and
  Vidal}]{chudasama2023enhancing}
Chudasama, Y.; Purohit, D.; Rohde, P.~D.; and Vidal, M.-E. 2023.
\newblock Enhancing Interpretability of Machine Learning Models over Knowledge
  Graphs.
\newblock In \emph{SEMANTiCS (Posters \& Demos)}.

\bibitem[{Clarke and Braun(2017)}]{clarke2017thematic}
Clarke, V.; and Braun, V. 2017.
\newblock Thematic analysis.
\newblock \emph{The journal of positive psychology}, 12(3): 297--298.

\bibitem[{Commission(2024)}]{EU_AI_Act_2024}
Commission, E. 2024.
\newblock Regulation (EU) 2024/1689 of the European Parliament and of the
  Council of 13 June 2024 laying down harmonised rules on artificial
  intelligence (Artificial Intelligence Act) and amending certain Union
  legislative acts.
\newblock \url{https://eur-lex.europa.eu/eli/reg/2024/1689/oj}.
\newblock Accessed: 28 December 2024.

\bibitem[{Commission(2025)}]{GPAI2025}
Commission, E. 2025.
\newblock GPAI Model Documentation Form.
\newblock
  \url{https://digital-strategy.ec.europa.eu/en/policies/contents-code-gpai}.
\newblock Accessed: 29 July 2025.

\bibitem[{Congress(2021)}]{USBillS2551_2021}
Congress, U. 2021.
\newblock S.2551 - A bill to establish guidelines for the use of artificial
  intelligence in government and to ensure that such use is effective,
  efficient, ethical, and accountable.
\newblock Accessed: 20 January 2025.

\bibitem[{Corbett-Davies et~al.(2023)Corbett-Davies, Gaebler, Nilforoshan,
  Shroff, and Goel}]{corbett2023measure}
Corbett-Davies, S.; Gaebler, J.~D.; Nilforoshan, H.; Shroff, R.; and Goel, S.
  2023.
\newblock The measure and mismeasure of fairness.
\newblock \emph{The Journal of Machine Learning Research}, 24(1): 14730--14846.

\bibitem[{Crisan et~al.(2022)Crisan, Drouhard, Vig, and
  Rajani}]{crisan2022interactive}
Crisan, A.; Drouhard, M.; Vig, J.; and Rajani, N. 2022.
\newblock Interactive model cards: A human-centered approach to model
  documentation.
\newblock In \emph{Proceedings of the 2022 ACM Conference on Fairness,
  Accountability, and Transparency}, 427--439.

\bibitem[{Critch and Russell(2023)}]{critch2023tasra}
Critch, A.; and Russell, S. 2023.
\newblock TASRA: a taxonomy and analysis of societal-scale risks from AI.
\newblock \emph{arXiv preprint arXiv:2306.06924}.

\bibitem[{D{\'\i}az et~al.(2022)D{\'\i}az, Kivlichan, Rosen, Baker, Amironesei,
  Prabhakaran, and Denton}]{diaz2022crowdworksheets}
D{\'\i}az, M.; Kivlichan, I.; Rosen, R.; Baker, D.; Amironesei, R.;
  Prabhakaran, V.; and Denton, E. 2022.
\newblock Crowdworksheets: Accounting for individual and collective identities
  underlying crowdsourced dataset annotation.
\newblock In \emph{Proceedings of the 2022 ACM Conference on Fairness,
  Accountability, and Transparency}, 2342--2351.

\bibitem[{D{\'\i}az-Rodr{\'\i}guez et~al.(2023)D{\'\i}az-Rodr{\'\i}guez,
  Del~Ser, Coeckelbergh, de~Prado, Herrera-Viedma, and
  Herrera}]{diaz2023connecting}
D{\'\i}az-Rodr{\'\i}guez, N.; Del~Ser, J.; Coeckelbergh, M.; de~Prado, M.~L.;
  Herrera-Viedma, E.; and Herrera, F. 2023.
\newblock Connecting the dots in trustworthy Artificial Intelligence: From AI
  principles, ethics, and key requirements to responsible AI systems and
  regulation.
\newblock \emph{Information Fusion}, 99: 101896.

\bibitem[{Digital Society~Initiative(2021)}]{SwissAIPositionPaper2021}
Digital Society~Initiative, U. o.~Z. 2021.
\newblock Position Paper: AI Regulation in Switzerland.
\newblock Accessed: 20 January 2025.

\bibitem[{Dobbe(2022)}]{dobbe2022system}
Dobbe, R. 2022.
\newblock System safety and artificial intelligence.
\newblock In \emph{Proceedings of the 2022 ACM Conference on Fairness,
  Accountability, and Transparency}, 1584--1584.

\bibitem[{for Security and (CSET)(2021)}]{ChinaAINorms2021}
for Security, C.; and (CSET), E.~T. 2021.
\newblock Ethical Norms for New Generation Artificial Intelligence Released.
\newblock Accessed: 20 January 2025.

\bibitem[{Force(2018)}]{NIST_SP_800-37r2}
Force, J.~T. 2018.
\newblock Risk Management Framework for Information Systems and Organizations:
  A System Life Cycle Approach for Security and Privacy.
\newblock Technical Report Special Publication (SP) 800-37, Rev. 2, National
  Institute of Standards and Technology, Gaithersburg, MD.

\bibitem[{Friedman and Nissenbaum(1996)}]{friedman1996bias}
Friedman, B.; and Nissenbaum, H. 1996.
\newblock Bias in computer systems.
\newblock \emph{ACM Transactions on information systems (TOIS)}, 14(3):
  330--347.

\bibitem[{Gebru et~al.(2021)Gebru, Morgenstern, Vecchione, Vaughan, Wallach,
  Iii, and Crawford}]{gebru_datasheets_2021}
Gebru, T.; Morgenstern, J.; Vecchione, B.; Vaughan, J.~W.; Wallach, H.; Iii,
  H.~D.; and Crawford, K. 2021.
\newblock Datasheets for datasets.
\newblock \emph{Communications of the ACM}, 64(12): 86--92.

\bibitem[{Gei and Jonsson(2023)}]{etami2023}
Gei, C.; and Jonsson, H. 2023.
\newblock etami: The Open Guidebook on Legal, Trustworthy, and Ethical
  Artificial Intelligence.
\newblock \url{https://guidebook.etami.org}.
\newblock Accessed: 8 August 2025.

\bibitem[{Golpayegani et~al.(2024)Golpayegani, Hupont, Panigutti, Pandit,
  Schade, O’Sullivan, and Lewis}]{golpayegani2024ai}
Golpayegani, D.; Hupont, I.; Panigutti, C.; Pandit, H.~J.; Schade, S.;
  O’Sullivan, D.; and Lewis, D. 2024.
\newblock AI cards: towards an applied framework for machine-readable AI and
  risk documentation inspired by the EU AI Act.
\newblock In \emph{Annual Privacy Forum}, 48--72. Springer.

\bibitem[{Gong et~al.(2023)Gong, Zhang, Wei, Zhang, and
  Huang}]{gong2023intended}
Gong, L.; Zhang, J.; Wei, M.; Zhang, H.; and Huang, Z. 2023.
\newblock What is the intended usage context of this model? An exploratory
  study of pre-trained models on various model repositories.
\newblock \emph{ACM Transactions on Software Engineering and Methodology},
  32(3): 1--57.

\bibitem[{Heger et~al.(2022)Heger, Marquis, Vorvoreanu, Wallach, and
  Wortman~Vaughan}]{heger2022understanding}
Heger, A.~K.; Marquis, L.~B.; Vorvoreanu, M.; Wallach, H.; and Wortman~Vaughan,
  J. 2022.
\newblock Understanding machine learning practitioners' data documentation
  perceptions, needs, challenges, and desiderata.
\newblock \emph{Proceedings of the ACM on Human-Computer Interaction},
  6(CSCW2): 1--29.

\bibitem[{Hiniduma et~al.(2024)Hiniduma, Byna, Bez, and
  Madduri}]{hiniduma2024ai}
Hiniduma, K.; Byna, S.; Bez, J.~L.; and Madduri, R. 2024.
\newblock AI data readiness inspector (AIDRIN) for quantitative assessment of
  data readiness for AI.
\newblock In \emph{Proceedings of the 36th International Conference on
  Scientific and Statistical Database Management}, 1--12.

\bibitem[{Holland et~al.(2020)Holland, Hosny, Newman, Joseph, and
  Chmielinski}]{holland2020dataset}
Holland, S.; Hosny, A.; Newman, S.; Joseph, J.; and Chmielinski, K. 2020.
\newblock The dataset nutrition label.
\newblock \emph{Data Protection and Privacy}, 12(12): 1.

\bibitem[{Hupont et~al.(2024)Hupont, Fern{\'a}ndez-Llorca, Baldassarri, and
  G{\'o}mez}]{hupont2024use}
Hupont, I.; Fern{\'a}ndez-Llorca, D.; Baldassarri, S.; and G{\'o}mez, E. 2024.
\newblock Use case cards: a use case reporting framework inspired by the
  European AI Act.
\newblock \emph{Ethics and Information Technology}, 26(2): 19.

\bibitem[{Hupont et~al.(2023)Hupont, Micheli, Delipetrev, G{\'o}mez, and
  Garrido}]{hupont2023documenting}
Hupont, I.; Micheli, M.; Delipetrev, B.; G{\'o}mez, E.; and Garrido, J.~S.
  2023.
\newblock Documenting high-risk AI: a European regulatory perspective.
\newblock \emph{Computer}, 56(5): 18--27.

\bibitem[{Hutchinson et~al.(2021)Hutchinson, Smart, Hanna, Denton, Greer,
  Kjartansson, Barnes, and Mitchell}]{hutchinson2021towards}
Hutchinson, B.; Smart, A.; Hanna, A.; Denton, E.; Greer, C.; Kjartansson, O.;
  Barnes, P.; and Mitchell, M. 2021.
\newblock Towards accountability for machine learning datasets: Practices from
  software engineering and infrastructure.
\newblock In \emph{Proceedings of the 2021 ACM Conference on Fairness,
  Accountability, and Transparency}, 560--575.

\bibitem[{Kale et~al.(2023)Kale, Nguyen, Harris~Jr, Li, Zhang, and
  Ma}]{kale2023provenance}
Kale, A.; Nguyen, T.; Harris~Jr, F.~C.; Li, C.; Zhang, J.; and Ma, X. 2023.
\newblock Provenance documentation to enable explainable and trustworthy AI: A
  literature review.
\newblock \emph{Data Intelligence}, 5(1): 139--162.

\bibitem[{Kim, Comuzzi, and Di~Francescomarino(2024)}]{kim2024explaining}
Kim, S.; Comuzzi, M.; and Di~Francescomarino, C. 2024.
\newblock Explaining the impact of design choices on model quality in
  predictive process monitoring.
\newblock \emph{Journal of Intelligent Information Systems}, 1--26.

\bibitem[{K{\"o}nigstorfer and Thalmann(2022)}]{konigstorfer2022ai}
K{\"o}nigstorfer, F.; and Thalmann, S. 2022.
\newblock AI Documentation: A path to accountability.
\newblock \emph{Journal of Responsible Technology}, 11: 100043.

\bibitem[{Kreuzberger, K{\"u}hl, and Hirschl(2023)}]{kreuzberger2023machine}
Kreuzberger, D.; K{\"u}hl, N.; and Hirschl, S. 2023.
\newblock Machine learning operations (mlops): Overview, definition, and
  architecture.
\newblock \emph{IEEE access}, 11: 31866--31879.

\bibitem[{Kroll(2021)}]{kroll2021outlining}
Kroll, J.~A. 2021.
\newblock Outlining traceability: A principle for operationalizing
  accountability in computing systems.
\newblock In \emph{Proceedings of the 2021 ACM Conference on Fairness,
  Accountability, and Transparency}, 758--771.

\bibitem[{Lee(2023)}]{lee2023collections}
Lee, B. C.~G. 2023.
\newblock The “Collections as ML Data” checklist for machine learning and
  cultural heritage.
\newblock \emph{Journal of the Association for Information Science and
  Technology}.

\bibitem[{Li et~al.(2023)Li, Qi, Liu, Di, Liu, Pei, Yi, and
  Zhou}]{li2023trustworthy}
Li, B.; Qi, P.; Liu, B.; Di, S.; Liu, J.; Pei, J.; Yi, J.; and Zhou, B. 2023.
\newblock Trustworthy AI: From principles to practices.
\newblock \emph{ACM Computing Surveys}, 55(9): 1--46.

\bibitem[{Loosley et~al.(2023)Loosley, Seifoddini, Canopoli, and
  Zehlike}]{loosley2023body}
Loosley, A.; Seifoddini, A.; Canopoli, A.; and Zehlike, M. 2023.
\newblock Body Measurement Prediction Fairness.
\newblock In \emph{Proceedings of the 2nd European Workshop on Algorithmic
  Fairness}.

\bibitem[{Lucaj, van~der Smagt, and Benbouzid(2023)}]{lucaj2023ai}
Lucaj, L.; van~der Smagt, P.; and Benbouzid, D. 2023.
\newblock Ai regulation is (not) all you need.
\newblock In \emph{Proceedings of the 2023 ACM Conference on Fairness,
  Accountability, and Transparency}, 1267--1279.

\bibitem[{Madaio et~al.(2020)Madaio, Stark, Wortman~Vaughan, and
  Wallach}]{madaio2020co}
Madaio, M.~A.; Stark, L.; Wortman~Vaughan, J.; and Wallach, H. 2020.
\newblock Co-designing checklists to understand organizational challenges and
  opportunities around fairness in AI.
\newblock In \emph{Proceedings of the 2020 CHI conference on human factors in
  computing systems}, 1--14.

\bibitem[{Marone and Van~Durme(2024)}]{marone2024data}
Marone, M.; and Van~Durme, B. 2024.
\newblock Data portraits: Recording foundation model training data.
\newblock \emph{Advances in Neural Information Processing Systems}, 36.

\bibitem[{Miceli et~al.(2021)Miceli, Yang, Naudts, Schuessler, Serbanescu, and
  Hanna}]{miceli2021documenting}
Miceli, M.; Yang, T.; Naudts, L.; Schuessler, M.; Serbanescu, D.; and Hanna, A.
  2021.
\newblock Documenting computer vision datasets: an invitation to reflexive data
  practices.
\newblock In \emph{Proceedings of the 2021 ACM Conference on Fairness,
  Accountability, and Transparency}, 161--172.

\bibitem[{Micheli et~al.(2023)Micheli, Hupont, Delipetrev, and
  Soler-Garrido}]{micheli2023landscape}
Micheli, M.; Hupont, I.; Delipetrev, B.; and Soler-Garrido, J. 2023.
\newblock The landscape of data and AI documentation approaches in the European
  policy context.
\newblock \emph{Ethics and Information Technology}, 25(4): 56.

\bibitem[{Mitchell et~al.(2019)Mitchell, Wu, Zaldivar, Barnes, Vasserman,
  Hutchinson, Spitzer, Raji, and Gebru}]{mitchell_model_2019}
Mitchell, M.; Wu, S.; Zaldivar, A.; Barnes, P.; Vasserman, L.; Hutchinson, B.;
  Spitzer, E.; Raji, I.~D.; and Gebru, T. 2019.
\newblock Model {Cards} for {Model} {Reporting}.
\newblock In \emph{Proceedings of the {Conference} on {Fairness},
  {Accountability}, and {Transparency}}, 220--229. Atlanta GA USA: ACM.
\newblock ISBN 978-1-4503-6125-5.

\bibitem[{Mittelstadt et~al.(2016)Mittelstadt, Allo, Taddeo, Wachter, and
  Floridi}]{mittelstadt2016ethics}
Mittelstadt, B.~D.; Allo, P.; Taddeo, M.; Wachter, S.; and Floridi, L. 2016.
\newblock The ethics of algorithms: Mapping the debate.
\newblock \emph{Big Data \& Society}, 3(2): 2053951716679679.

\bibitem[{Naderifar, Goli, and Ghaljaie(2017)}]{naderifar2017snowball}
Naderifar, M.; Goli, H.; and Ghaljaie, F. 2017.
\newblock Snowball sampling: A purposeful method of sampling in qualitative
  research.
\newblock \emph{Strides in development of medical education}, 14(3).

\bibitem[{Naja et~al.(2021)Naja, Markovic, Edwards, and
  Cottrill}]{naja2021semantic}
Naja, I.; Markovic, M.; Edwards, P.; and Cottrill, C. 2021.
\newblock A semantic framework to support AI system accountability and audit.
\newblock In \emph{The Semantic Web: 18th International Conference, ESWC 2021,
  Virtual Event, June 6--10, 2021, Proceedings 18}, 160--176. Springer.

\bibitem[{Nodes(2024)}]{UKAIRegulations2024}
Nodes, L. 2024.
\newblock UK AI Regulations: An Overview.
\newblock Accessed: 20 January 2025.

\bibitem[{of~Science and Policy(2022)}]{USAIBillOfRights2022}
of~Science, T. W. H.~O.; and Policy, T. 2022.
\newblock Blueprint for an AI Bill of Rights: Making Automated Systems Work for
  the American People.
\newblock Accessed: 20 January 2025.

\bibitem[{O'neil(2017)}]{o2017weapons}
O'neil, C. 2017.
\newblock \emph{Weapons of math destruction: How big data increases inequality
  and threatens democracy}.
\newblock Crown.

\bibitem[{Papakyriakopoulos et~al.(2023)Papakyriakopoulos, Choi, Thong, Zhao,
  Andrews, Bourke, Xiang, and Koenecke}]{papakyriakopoulos2023augmented}
Papakyriakopoulos, O.; Choi, A. S.~G.; Thong, W.; Zhao, D.; Andrews, J.;
  Bourke, R.; Xiang, A.; and Koenecke, A. 2023.
\newblock Augmented datasheets for speech datasets and ethical decision-making.
\newblock In \emph{Proceedings of the 2023 ACM Conference on Fairness,
  Accountability, and Transparency}, 881--904.

\bibitem[{Partnership(2023)}]{BrazilAIBill2023}
Partnership, A. 2023.
\newblock Access Alert: Brazil’s New AI Bill – A Comprehensive Framework
  for Ethical and Responsible Use of AI Systems.
\newblock Accessed: 20 January 2025.

\bibitem[{Procope et~al.(2022)Procope, Cheema, Adkins, Alsallakh, Green,
  McReynolds, Pehl, Wang, and Zvyagina}]{procope2022system}
Procope, C.; Cheema, A.; Adkins, D.; Alsallakh, B.; Green, N.; McReynolds, E.;
  Pehl, G.; Wang, E.; and Zvyagina, P. 2022.
\newblock System-level transparency of machine Learning.
\newblock \emph{Meta Research. February}, 23.

\bibitem[{Project(2022)}]{ChinaAIRegulation2022}
Project, S.~D. 2022.
\newblock Translation: Internet Information Service Algorithmic Recommendation
  Management Provisions.
\newblock Accessed: 20 January 2025.

\bibitem[{Pushkarna, Zaldivar, and Kjartansson(2022)}]{pushkarna2022data}
Pushkarna, M.; Zaldivar, A.; and Kjartansson, O. 2022.
\newblock Data cards: Purposeful and transparent dataset documentation for
  responsible ai.
\newblock In \emph{Proceedings of the 2022 ACM Conference on Fairness,
  Accountability, and Transparency}, 1776--1826.

\bibitem[{Raji et~al.(2022)Raji, Kumar, Horowitz, and Selbst}]{raji2022fallacy}
Raji, I.~D.; Kumar, I.~E.; Horowitz, A.; and Selbst, A. 2022.
\newblock The fallacy of AI functionality.
\newblock In \emph{Proceedings of the 2022 ACM Conference on Fairness,
  Accountability, and Transparency}, 959--972.

\bibitem[{Raji et~al.(2020)Raji, Smart, White, Mitchell, Gebru, Hutchinson,
  Smith-Loud, Theron, and Barnes}]{raji2020closing}
Raji, I.~D.; Smart, A.; White, R.~N.; Mitchell, M.; Gebru, T.; Hutchinson, B.;
  Smith-Loud, J.; Theron, D.; and Barnes, P. 2020.
\newblock Closing the AI accountability gap: Defining an end-to-end framework
  for internal algorithmic auditing.
\newblock In \emph{Proceedings of the 2020 conference on fairness,
  accountability, and transparency}, 33--44.

\bibitem[{Register(2019)}]{USAILeadership2019}
Register, F. 2019.
\newblock Executive Order on Maintaining American Leadership in Artificial
  Intelligence.
\newblock Accessed: 20 January 2025.

\bibitem[{Register(2020)}]{USTrustworthyAI2020}
Register, F. 2020.
\newblock Executive Order on Promoting the Use of Trustworthy Artificial
  Intelligence in the Federal Government.
\newblock Accessed: 20 January 2025.

\bibitem[{Richards et~al.(2020)Richards, Piorkowski, Hind, Houde, and
  Mojsilovi{\'c}}]{richards2020methodology}
Richards, J.; Piorkowski, D.; Hind, M.; Houde, S.; and Mojsilovi{\'c}, A. 2020.
\newblock A methodology for creating AI FactSheets.
\newblock \emph{arXiv preprint arXiv:2006.13796}.

\bibitem[{Roman et~al.(2023)Roman, Vaughan, See, Ballard, Torres, Robinson, and
  Ferres}]{roman2023open}
Roman, A.~C.; Vaughan, J.~W.; See, V.; Ballard, S.; Torres, J.; Robinson, C.;
  and Ferres, J. M.~L. 2023.
\newblock Open datasheets: machine-readable documentation for open datasets and
  responsible AI assessments.
\newblock \emph{arXiv preprint arXiv:2312.06153}.

\bibitem[{Schoenherr et~al.(2023)Schoenherr, Abbas, Michael, Rivas, and
  Anderson}]{schoenherr2023designing}
Schoenherr, J.~R.; Abbas, R.; Michael, K.; Rivas, P.; and Anderson, T.~D. 2023.
\newblock Designing AI using a human-centered approach: Explainability and
  accuracy toward trustworthiness.
\newblock \emph{IEEE Transactions on Technology and Society}, 4(1): 9--23.

\bibitem[{Seifoddini et~al.(2023)Seifoddini, Vernooij, Künzle, Canopoli, Alf,
  Volokitin, and Shirvany}]{seifoddini2023alisnet}
Seifoddini, A.; Vernooij, K.; Künzle, T.; Canopoli, A.; Alf, M.; Volokitin,
  A.; and Shirvany, R. 2023.
\newblock ALiSNet: Accurate and Lightweight Human Segmentation Network for
  Fashion E-Commerce.
\newblock In \emph{Proceedings of the 18th International Joint Conference on
  Computer Vision, Imaging and Computer Graphics Theory and Applications
  (VISIGRAPP 2023) - Volume 4: VISAPP}, 746--754. INSTICC, SciTePress.
\newblock ISBN 978-989-758-634-7.

\bibitem[{Shen et~al.(2021)Shen, Deng, Chattopadhyay, Wu, Wang, and
  Zhu}]{shen2021value}
Shen, H.; Deng, W.~H.; Chattopadhyay, A.; Wu, Z.~S.; Wang, X.; and Zhu, H.
  2021.
\newblock Value cards: An educational toolkit for teaching social impacts of
  machine learning through deliberation.
\newblock In \emph{Proceedings of the 2021 ACM conference on fairness,
  accountability, and transparency}, 850--861.

\bibitem[{Shneiderman(2020)}]{shneiderman2020bridging}
Shneiderman, B. 2020.
\newblock Bridging the gap between ethics and practice: guidelines for
  reliable, safe, and trustworthy human-centered AI systems.
\newblock \emph{ACM Transactions on Interactive Intelligent Systems (TiiS)},
  10(4): 1--31.

\bibitem[{Smart et~al.(2024)Smart, Ahmed, Metcalf, Kasirzadeh, Belli, Rismani,
  Dobbe et~al.}]{smart2024sociotechnical}
Smart, A.; Ahmed, S.; Metcalf, J.; Kasirzadeh, A.; Belli, L.; Rismani, S.;
  Dobbe, R.; et~al. 2024.
\newblock What Is Sociotechnical AI Safety? What Do We Want It To Be? A FAccT
  Community Workshop.
\newblock Presented at the FAccT Community Workshop.

\bibitem[{Smuha(2021)}]{smuha2021beyond}
Smuha, N.~A. 2021.
\newblock Beyond the individual: governing AI’s societal harm.
\newblock \emph{Internet Policy Review}, 10(3).

\bibitem[{Sokol and Flach(2020)}]{sokol2020explainability}
Sokol, K.; and Flach, P. 2020.
\newblock Explainability fact sheets: a framework for systematic assessment of
  explainable approaches.
\newblock In \emph{Proceedings of the 2020 conference on fairness,
  accountability, and transparency}, 56--67.

\bibitem[{Solow-Niederman(2023)}]{solow2023algorithmic}
Solow-Niederman, A.~G. 2023.
\newblock Algorithmic grey holes.
\newblock \emph{JL \& Innovation}, 5: 116.

\bibitem[{Sovrano et~al.(2022)Sovrano, Sapienza, Palmirani, and
  Vitali}]{sovrano2022metrics}
Sovrano, F.; Sapienza, S.; Palmirani, M.; and Vitali, F. 2022.
\newblock Metrics, explainability and the European AI act proposal.
\newblock \emph{J}, 5(1): 126--138.

\bibitem[{Squires(2023)}]{squires2023thematic}
Squires, V. 2023.
\newblock Thematic analysis.
\newblock In \emph{Varieties of qualitative research methods: Selected
  contextual perspectives}, 463--468. Springer.

\bibitem[{Tagliabue et~al.(2021)Tagliabue, Tuulos, Greco, and
  Dave}]{tagliabue2021dag}
Tagliabue, J.; Tuulos, V.; Greco, C.; and Dave, V. 2021.
\newblock DAG Card is the new Model Card.
\newblock \emph{arXiv preprint arXiv:2110.13601}.

\bibitem[{Terry et~al.(2017)Terry, Hayfield, Clarke, Braun
  et~al.}]{terry2017thematic}
Terry, G.; Hayfield, N.; Clarke, V.; Braun, V.; et~al. 2017.
\newblock Thematic analysis.
\newblock \emph{The SAGE handbook of qualitative research in psychology},
  2(17-37): 25.

\bibitem[{Veale and Zuiderveen~Borgesius(2021)}]{veale2021demystifying}
Veale, M.; and Zuiderveen~Borgesius, F. 2021.
\newblock Demystifying the Draft EU Artificial Intelligence Act—Analysing the
  good, the bad, and the unclear elements of the proposed approach.
\newblock \emph{Computer Law Review International}, 22(4): 97--112.

\bibitem[{Winecoff and Bogen(2024)}]{winecoff2024improving}
Winecoff, A.~A.; and Bogen, M. 2024.
\newblock Improving governance outcomes through AI documentation: Bridging
  theory and practice.
\newblock \emph{arXiv preprint arXiv:2409.08960}.

\bibitem[{Zehlike et~al.(2025)Zehlike, Loosley, Jonsson, Wiedemann, and
  Hacker}]{zehlike2025beyond}
Zehlike, M.; Loosley, A.; Jonsson, H.; Wiedemann, E.; and Hacker, P. 2025.
\newblock Beyond incompatibility: Trade-offs between mutually exclusive
  fairness criteria in machine learning and law.
\newblock \emph{Artificial Intelligence}, 104280.

\end{thebibliography}

\end{document}